\documentclass[conference]{IEEEtran}
\IEEEoverridecommandlockouts
\usepackage{cite}
\usepackage{amsmath,amssymb,amsfonts}
\usepackage{algorithmic}
\usepackage{graphicx}
\usepackage{textcomp}
\usepackage[numbers]{natbib}
\usepackage{booktabs}
\usepackage{xcolor}
\usepackage{hyperref}

\def\BibTeX{{\rm B\kern-.05em{\sc i\kern-.025em b}\kern-.08em
    T\kern-.1667em\lower.7ex\hbox{E}\kern-.125emX}}
\begin{document}

\title{Predicting the Skies: A Novel Model for Flight-Level Passenger Traffic Forecasting
% \\
}

% \author{\IEEEauthorblockN{1\textsuperscript{st} Sina Ehsani}
\author{\IEEEauthorblockN{Sina Ehsani}
\IEEEauthorblockA{\textit{Industrial Engineering} \\
\textit{University of Arizona}\\
sinaehsani@arizona.edu}
\and
\IEEEauthorblockN{Elina Sergeeva}
\IEEEauthorblockA{\textit{Operations Research} \\
\textit{American Airlines}\\
elina.sergeeva@aa.com}
\and
\IEEEauthorblockN{Wendy Murdy}
\IEEEauthorblockA{\textit{Operations Research} \\
\textit{American Airlines}\\
wendy.murdy@aa.com}
\and
\IEEEauthorblockN{Benjamin Fox}
\IEEEauthorblockA{\textit{Operations Research} \\
\textit{American Airlines}\\
benjamin.fox1@aa.com}
% \and
% \and
% \and
}

\maketitle

\begin{abstract}
Accurate prediction of flight-level passenger traffic is of paramount importance in airline operations, influencing key decisions from pricing to route optimization. This study introduces a novel, multimodal deep learning approach to the challenge of predicting flight-level passenger traffic, yielding substantial accuracy improvements compared to traditional models. Leveraging an extensive dataset from American Airlines, our model ingests historical traffic data, fare closure information, and seasonality attributes specific to each flight. Our proposed neural network integrates the strengths of Recurrent Neural Networks (RNN) and Convolutional Neural Networks (CNN), exploiting the temporal patterns and spatial relationships within the data to enhance prediction performance. Crucial to the success of our model is a comprehensive data processing strategy. We construct 3D tensors to represent data, apply careful masking strategies to mirror real-world dynamics, and employ data augmentation techniques to enrich the diversity of our training set. The efficacy of our approach is borne out in the results: our model demonstrates an approximate 33\% improvement in Mean Squared Error (MSE) compared to traditional benchmarks. This study, therefore, highlights the significant potential of deep learning techniques and meticulous data processing in advancing the field of flight traffic prediction.
\end{abstract}

\begin{IEEEkeywords}
Multimodal Deep Learning, Spatial and Sequential Relations, Traffic Prediction, Airline Industry, Machine Learning
\end{IEEEkeywords}
\section{Introduction}

Flight traffic prediction plays a critical role in the strategic and operational management of airlines, directly affecting key decisions such as pricing and route planning \citep{mayer2003network, zachariah2023systematic}. Traditional statistical models, while useful, often fail to capture the multifaceted nature of passenger traffic, which is influenced by a complex interplay of factors  \citep{hewamalage2021recurrent,al2016forecasting}. Recognizing this gap, our study proposes a deep learning framework that is specifically designed to grapple with the intricacies of airline data, providing a more holistic and accurate prediction method.

In this study, we introduce an advanced deep learning model that utilizes an extensive dataset from American Airlines. This dataset encompasses not only historical passenger traffic data but also detailed seasonality attributes—such as peak travel times and holiday effects—and dynamic fare closure information, which includes the timing and patterns of fare availability changes. By incorporating this multifaceted dataset into our deep learning model, we aim to significantly enhance the accuracy of flight traffic predictions \citep{dobruszkes2022monthly}.

Our study contributes to the field in several ways:

\begin{itemize}
\item We introduce a 3D data transformation technique, aiming to make better use of past traffic, seasonality information, and fare closure data for prediction purposes \citep{lecun2015deep}.
\item We explore a multimodal deep learning approach designed to detect both spatial and sequential patterns in the data. This method seeks to build upon previous methodologies and potentially enhance prediction performance.
\item We employ innovative masking and data augmentation strategies for structured data \citep{wong2016understanding}.
\item Our model's performance is compared with prior statistical and machine learning methods, showing promise in terms of detecting recent trends and adhering to main seasonality patterns.
\end{itemize}

The overarching goal of our research is to leverage our deep learning model to simulate a variety of fare closure scenarios, thereby gaining insights into passenger booking behaviors. Accurately predicting how different fare restrictions affect bookings allows us to guide airlines in crafting fare strategies that optimize both profitability and customer satisfaction \citep{you1999dynamic}. This predictive capability represents a significant stride towards more intelligent and adaptive airline revenue management \citep{tian2021data}.

The remainder of this paper is methodically structured to guide the reader through our research process. We begin with a review of related work that traces the evolution from traditional airline traffic prediction methods to cutting-edge machine learning techniques. Subsequently, we delve into our dataset and the meticulous preprocessing methods employed, including our unique 3D tensor transformation and the strategic application of masking. The description of our multimodal deep learning model follows, elucidating its architecture and the rationale behind its design. We then present the experimental results, demonstrating our model's superior performance against established benchmarks. The paper concludes by reflecting on the limitations of our current model and proposing avenues for future inquiry.

\section{Related Work}

Forecasting passenger flight-level demand continues to be a critical component in airline revenue management. Traditional methods, including moving averages, exponential smoothing, and regression models, have established the foundation for such predictions \citep{hyndman2018forecasting, box2015time}. Additionally, time series analysis using ARIMA and SARIMA models has been pivotal in advancing these forecasting techniques, showcasing their effectiveness in capturing more nuanced patterns in flight demand \citep{carmona2020sarima, do2020forecasting}.

Dynamic Modeling Techniques have increasingly been recognized for their adaptability to the volatile and irregular nature of air traffic demand. These techniques, including hybrid approaches like seasonal decomposition and support vector regression, have proven effective in long-term forecasting, accurately capturing the fluctuating patterns of airline traffic \citep{chudy2017seasonal, xie2014short}. A landmark development in dynamic forecasting was American Airlines' DINAMO program, which smartly adjusted fares based on predicted versus actual bookings \citep{donovan2005yield}. The evolution of time series analysis has made it a mainstay in forecasting, with recent advancements demonstrating its superiority over some statistical methods when applied to specific flight demand profile \citep{chen2012improving}. Concurrently, econometric models, utilizing broader economic and demographic datasets, have emerged as an alternative approach, enriching the landscape of demand forecasting \citep{wooldridge2015introductory}

The advent of Big Data, coupled with the rise of machine learning, has significantly transformed forecasting methodologies. Techniques that utilize big data have unlocked the potential to uncover complex patterns and trends, offering deeper insights into passenger behavior and preferences that surpass the capabilities of traditional models \citep{li2020forecasting, kim2016forecasting}.

In parallel, the emergence of neural network has marked a new era in forecasting. These networks provide considerable advantages over traditional time-series and econometric models, demonstrating enhanced accuracy and predictive power \citep{zhang1998forecasting, cicek2021optimizing}. Recent advancements further underscore the effectiveness of neural networks in forecasting, with specific models like ConvLSTM gaining prominence for their ability to capture spatial-temporal patterns, crucial for demand forecasting in the aviation industry \citep{shi2015convolutional, ke2017short, muros2022air}.

Building upon the foundational advancements in neural networks and big data, recent studies have ventured into even more specialized and innovative methods. The Temporal Fusion Transformer (TFT) model, for instance, has shown promising results in predicting strategic flight departure demand across various airports and time horizons, showcasing its versatility \citep{wang2022flight}. Similarly, the multi-task adaptive graph attention network represents a leap forward in region-level travel demand forecasting, pushing the boundaries of granularity in demand prediction \citep{liang2023region}. 

In tandem with these advancements, machine learning has been increasingly applied to closely related areas such as flight fare prediction, acknowledging its growing significance in the digital market \citep{alapati2022prediction}. Moreover, ensemble learning methods have emerged as vital tools in temporal-spatial resource optimization, addressing the escalating challenges in managing air traffic demand \citep{xu2023data}.

Our research advances the field of airline passenger traffic forecasting by integrating deep learning approaches, such as ConvLSTM \citep{shi2015convolutional} and the DeepShallow network \citep{ehsani2023deepshallow}. This integration not only enhances the accuracy in predicting passenger traffic but also provides nuanced insights into complex booking behaviors influenced by fare closure scenarios. Significantly, our model's adaptability, as demonstrated in the \autoref{Sensitivity_Analysis}, is particularly relevant in the context of recent challenges like pandemics and other external factors. It showcases the ability to swiftly adjust to unforeseen market changes, highlighting the critical need for flexible and dynamic tools in airline revenue management \citep{gossling2020pandemics}. Such adaptability underscores our model's potential as a sophisticated analytical tool, equipped to handle the unpredictability inherent in modern airline operations.

% ---------------------************-------------------------
% ----------------------Dataset-------------------------
% ---------------------************-------------------------

\section{Dataset}

Our analysis is underpinned by a rich dataset divided into three essential components: seasonality features, historical traffic data, and fare closure information. Each segment provides a distinct perspective and detailed insight into flight operations, which collectively enhance the predictive capabilities of our model.

\subsection{Seasonality Features}

The seasonality features of our dataset encompass a variety of factors that exhibit cyclical trends and influence passenger demand. These features include temporal aspects such as the day of the week, week of the year, and holiday occurrences, as well as specific flight details like origin and destination airports, available seating capacity, and the historical Revenue per Available Seat Mile (RASM). By integrating these seasonality features with historical traffic data, we enrich our model's ability to understand and predict flow of passenger traffic, taking into account both regular and exceptional patterns that affect flight bookings.

\subsection{Historical Traffic}

Our dataset includes historical traffic data that details the number of passengers booked on each flight, segmented by specific periods relative to the flight's departure (e.g., 1-5 days before, 6-10 days before, etc.) and by fare class brackets (e.g., between \$0-\$100, \$100-\$200, etc.).  Additionally, the data distinguishes between local and flow traffic, offering insights into direct versus connecting passenger booking behaviors. This segmented historical analysis is instrumental in revealing trends and patterns within passenger demand that our model can learn from.

\subsection{Fare Closure}

Fare closure refers to specific time ranges during which certain fare classes or prices are made unavailable for booking. This could be in response to predicted demand, seat availability, or strategic pricing decisions by the airline. By restricting access to certain fares, airlines can optimize revenue and manage passenger loads more effectively. In our dataset, fare closure data provides insights into the availability of these fare classes over time. For each segment similar to those used for historical traffic data, it indicates the average openness or closeness of a fare class. If a fare was closed for a certain time period, passengers would be unable to book a ticket at that price. By incorporating this data, we acknowledge the dynamic nature of fare availability and its potential impact on passenger demand. This feature enables our model to capture the influence of price fluctuations and restrictions on the booking patterns.

Collectively, these data components provide a comprehensive view of the factors influencing flight-level passenger demand. The richness and granularity of the dataset form the bedrock of our neural network model, allowing it to effectively learn and forecast future passenger traffic. In the \autoref{sub:dataprocessing} we will go over the preprocessing of the data.

% ---------------------************-------------------------
% ----------------------Baselines-------------------------
% ---------------------************-------------------------

\section{Baselines}

This study relies on several established statistical and deep learning models as baselines to forecast future flight-level passenger demand. Each model utilized a combination of historical traffic, fare closure data, and seasonality factors. Our selection includes traditional models like Autoregressive Integrated Moving Average (ARIMA) and Seasonal ARIMA (SARIMA), alongside machine learning methods.

\subsubsection*{Statistical Models}

ARIMA and SARIMA serve as two of our primary statistical models. As a prevalent time-series forecasting method, ARIMA is equipped to handle univariate data exhibiting non-stationarity. This model is fitted to time-series data to dissect underlying patterns and generate reliable forecasts. SARIMA extends ARIMA's functionality, incorporating seasonality to better capture linear and seasonal trends in data. Given the inherent seasonal patterns in flight-level passenger demand, SARIMA proves particularly applicable.

\subsubsection*{Machine Learning Models}

For our baseline models, we focus on Convolutional Neural Networks (CNNs). CNNs are employed to extract spatial features from the data. In our implementation, they primarily concentrate on same day fare closure and seasonality data, using these factors to predict traffic for any given day. Their ability to effectively capture spatial hierarchies in data makes them a preferred choice for this type of information.

Transitioning to time-series data, we utilize the capabilities of the Convolutional LSTM (ConvLSTM) networks. The ConvLSTM, a hybrid model combining the spatial feature extraction properties of CNNs with the sequence modeling capabilities of Long Short-Term Memory (LSTM) networks, is particularly suited for our task. It can effectively use historical traffic data, taking into account both spatial and temporal dependencies, to predict future traffic. By comparing our proposed method against these baseline models, we aim to demonstrate the relative efficacy of our model in predicting flight-level passenger demand.

% ---------------------************-------------------------
% ----------------------Approach-------------------------
% ---------------------************-------------------------

\section{Approach}

The approach we adopted to predict future flight-level passenger demand necessitated a combination of diligent data processing and an innovative model architecture, each playing a significant role in the success of our methodology. This section will describe the components of our strategy, outlining our methods for data processing and the structure of our model.

\subsection{Data Processing}
\label{sub:dataprocessing}

\begin{figure}[h!]
    \centering
 \includegraphics[width=\linewidth]
    {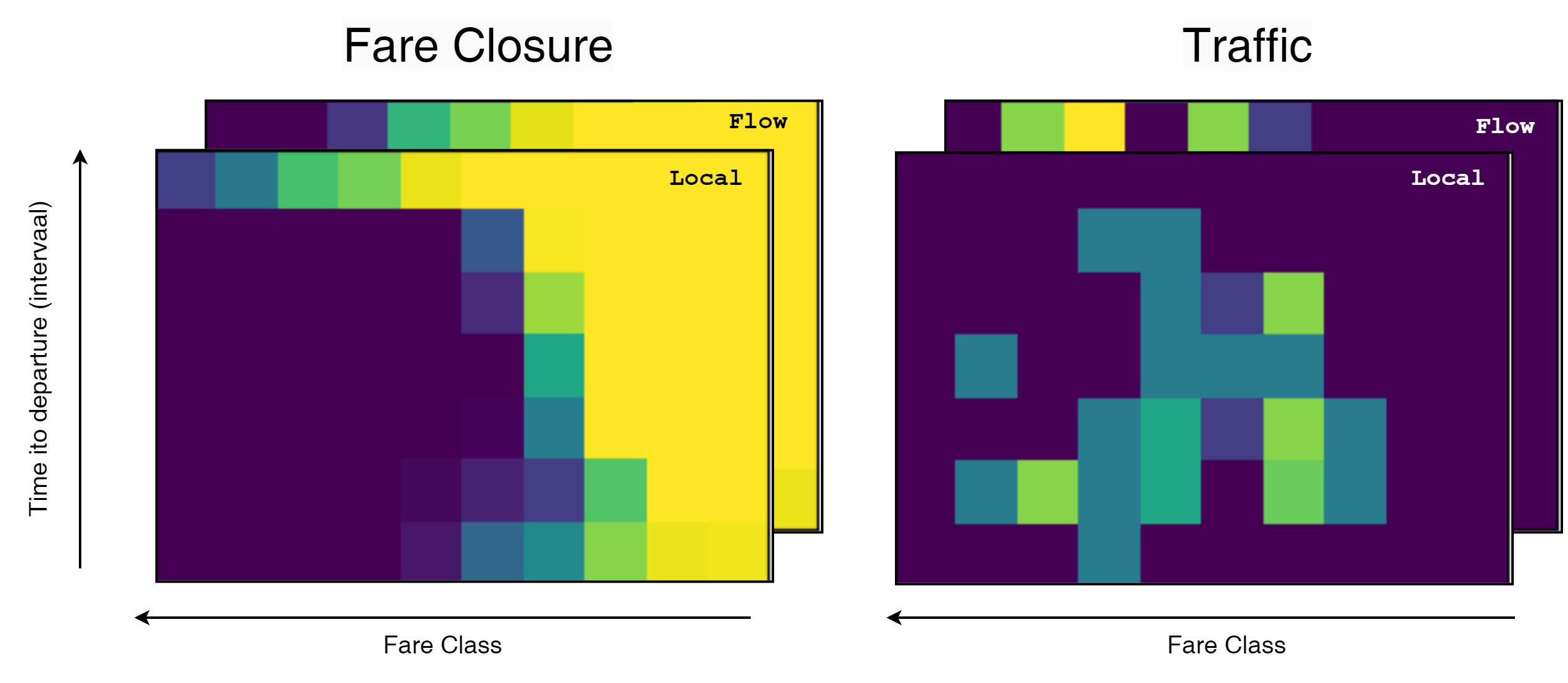}
    \caption{Figure X: 3D Tensor Visualization of Fare Closure and Traffic Data for a Specific Flight on a Given Date. This matrix represents a three-dimensional dataset where the x-axis delineates different fare classes, the y-axis corresponds to time to departure (measured in intervals), and the z-axis differentiates between traffic types—local versus flow. The fare closure data is normalized between 0 and 1 for any market, indicating the availability of fare classes over time, with 1 being full closure. Traffic data is scaled by the number of passengers booked within each price range, with the minimum set at 0. The color gradient represents these quantities, with purple signifying the minimum and yellow indicating the maximum value. This visual encoding is designed to reveal the dynamic relationship between fare availability, passenger booking behavior, and traffic type as the departure time approaches.}
    \label{fig:3d_data}
\end{figure}

The preprocessing of our dataset plays a crucial role in shaping the data to fit our neural network model. It consists of four essential steps: creating 3D tensors to capture spatial relationships, incorporating historical traffic data, designing a masking strategy for training, validation, and test sets, and employing data augmentation techniques to enhance the training process. Additionally, we ensure that all the data is normalized to aid in the learning process.

\subsubsection{Creating 3D Tensors Capturing Spatial Relations}

\begin{figure*}[h!]
    \centering
 \includegraphics[width=\textwidth]
    {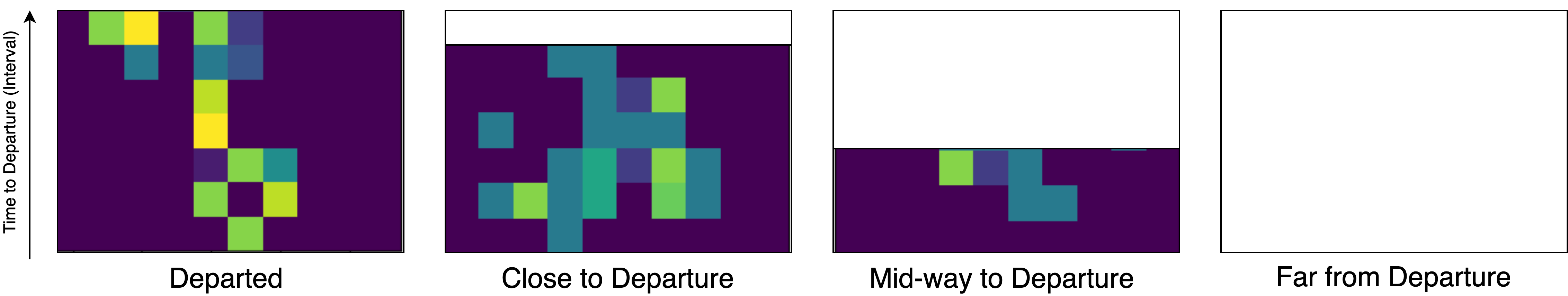}
    \caption{Data Completeness Across Different Timeframes to Departure. This figure highlights the progression of data availability as it correlates with the time remaining until a flight's departure. The visual is segmented into four panels, each representing a different stage relative to the departure time: 'Departed', 'Close to Departure', 'Mid-way to Departure', and 'Far from Departure'. Within each panel, the x-axis categorizes fare classes, while the y-axis measures time to departure in intervals. The color coding is indicative of booking data, where purple signifies no bookings (value 0), yellow indicates a high volume of bookings, and white represents missing data due to the information not being available or the flight being in the future. The contrast between panels clearly shows that data for flights 'Far from Departure' have the most white spaces, reflecting a high degree of incompleteness, which progressively diminishes as flights near the departure date, culminating in the 'Departed' panel, where the dataset is fully detailed with booking patterns.}
    \label{fig:incomplete_data}
\end{figure*}

The passenger traffic data and fare closure information for each flight are encoded as 3D tensors. The tensor axes correspond to fare class and time-to-departure, while the third axis differentiates between local and flow traffic types. As depicted in \autoref{fig:3d_data}, this multidimensional representation maintains the spatial relationships inherent in the data, facilitating a clear visualization of the interplay between fare closure rates and traffic patterns. Nevertheless, as demonstrated in  \autoref{fig:incomplete_data}, the integrity of our dataset is contingent upon the temporal proximity to the flight's departure date. For flights that have already transpired, our dataset is complete; however, for imminent flights, specific details have yet to materialize. The further a flight is from its departure, the greater the number of time intervals for which data remains uncollected. Such variability has necessitated the development of novel methodologies to mitigate the impact of these data gaps.

\subsubsection{Leveraging Historical Traffic Patterns for Enhanced Traffic Forecasting}

Our traffic prediction model is designed to utilize historical traffic data specific to each flight. In-house analysis has revealed a pronounced correlation between the traffic patterns of a flight and those from the same weekday in the preceding week, rather than the immediately previous day. For instance, the passenger volume of a flight on a Monday is more predictive of the traffic from the Monday of the prior week than that of the previous Sunday. To capitalize on this weekly cyclical pattern, our forecasting incorporates traffic data from the past n flights that occurred on the same day of the week. By adopting this window-size approach, we effectively harness recurring traffic dynamics, thereby refining the accuracy of our traffic predictions.

\subsubsection{Masking Protocols for Dataset Segmentation}

In the process of partitioning our dataset for model training and validation, we have adopted a dual approach that maintains the temporal sequence of data while introducing a novel masking technique. The most current three-month segment of the dataset is designated as the test set to ensure the model's performance is gauged against the most recent data. The remaining data is distributed into training and validation sets in a 90\% to 10\% split, adhering to the original chronological sequence.
For the validation and test datasets, we employ a specific reference point—the start date of each respective subset. In relation to this date, we apply a masking operation to the traffic tensors for instances that pertain to forthcoming, not yet realized fare closures, by assigning them a value of -1. This procedure is particularly applied to the validation and test sets, facilitating the model's ability to deduce traffic trends through the observed patterns in fare closures.
In a different vein, the training dataset is subjected to a variable masking technique to simulate a range of temporal conditions. Each data point is masked based on a randomly determined departure date. This method introduces diverse temporal scenarios into the training dataset, thereby broadening the scope of potential booking situations the model encounters and enhancing the robustness of the training regimen.

Given a tensor \(T\) comprising elements \( t_{ijk} \), where \( i \), \( j \), and \( k \) denote the indices for fare class, time to departure, and traffic type, respectively, we define the masking operation for the training set as follows:

\begin{equation}
t_{ijk}^{\prime} = 
\begin{cases} 
-1, & \text{if } j > J_{random} \\
t_{ijk}, & \text{otherwise}
\end{cases}
\end{equation}

where \( J_{random} \) represents the randomly selected index for time to departure, corresponding to the assigned pseudo-random departure date.

\subsubsection{Data Augmentation Techniques}

The data augmentation methodology employed in our study builds upon the masking techniques previously outlined for the training set. To foster the model's generalization across varied scenarios, the departure dates for individual data points are randomized subsequent to each training epoch. This randomization induces a corresponding adjustment in the masking of data points, as they align with their new departure dates.
As a result, the training dataset undergoes a transformation with each epoch, introducing the model to a fresh array of masked data arrangements. This dynamic process of consistent reshuffling and reapplication of masks ensures that the model is trained on a diverse set of data scenarios throughout its learning phases.

\subsubsection{Feature Normalization in Data Preprocessing}

In our preprocessing pipeline, we have implemented a feature normalization protocol alongside our data augmentation techniques. By uniformly scaling all input features, this process is crucial in maintaining balance in the model's performance, ensuring that no feature disproportionately influences the learning phase due to its scale. This step not only eliminates potential biases but also ensures that all features contribute equally to the predictive outcomes of our network model \citep{qi2022note}. Such normalization is key to creating an equitable learning environment, crucial for the effectiveness of our model.

\subsection{Architecture of the Model}

\begin{figure*}[h!]
    \centering
 \includegraphics[width=\textwidth]
    {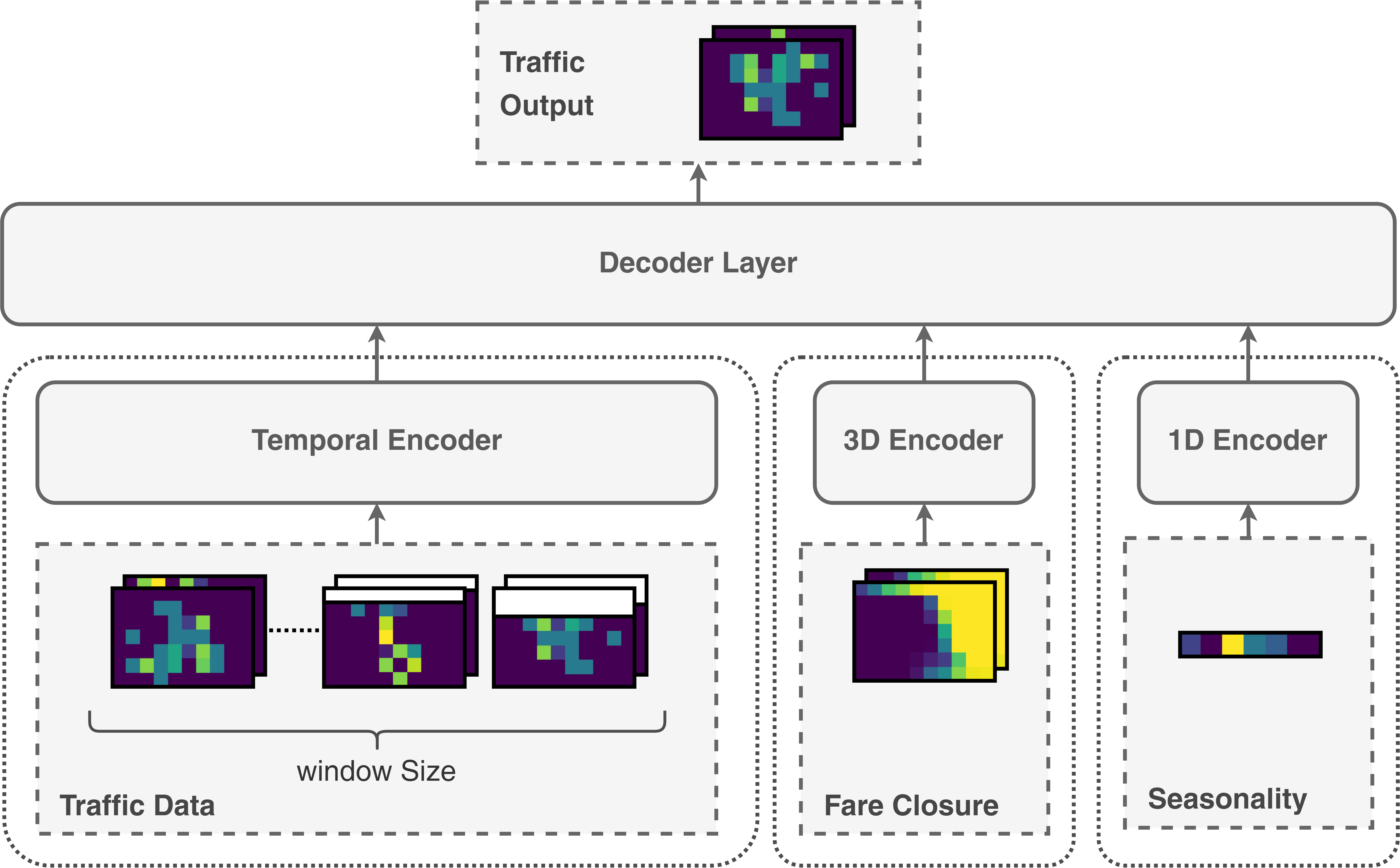}
    \caption{Schematic of the Model Architecture for Flight-Level Passenger Traffic Prediction. This figure delineates the configuration of the model's three encoders and the subsequent decoder layer. The Temporal Encoder processes time series traffic data, transforming it through a sequence of operations to capture temporal dynamics. The 3D Encoder manages the spatial aspects of fare closure data, employing volumetric filters to comprehend price-related variations. The 1D Encoder is specialized for discerning seasonality patterns, ensuring cyclical trends are captured. These processed inputs are then synthesized in the Decoder layer through a series of CNN layers that integrate the temporal, spatial, and seasonal features. The model's architecture is designed to preserve and utilize spatial relationships, culminating in a passenger traffic tensor that reflects a spatially informed forecast. The visualization provides insight into the input and output data shapes, illustrating the comprehensive flow from raw data to traffic prediction.}
    \label{fig:modeloverview}
\end{figure*}

'The model introduced in this research is structured to forecast passenger traffic on a per-flight basis. It integrates an analysis of historical traffic data (\( T_h \)), fare closure information (\( F_c \)), and seasonality factors (\( S_f \)). The model’s architecture, comprising three encoders and a decoder, is formulated to capture the spatial-temporal dynamics inherent in these data streams. A schematic overview is presented in \autoref{fig:modeloverview}.

\subsubsection{Temporal Dependency Encoder}

The temporal dependency encoder (\( E_{td} \)) is designed to encapsulate temporal patterns in traffic data. Experimentation with ConvLSTM \citep{shi2015convolutional} and the DeepShallow Network layer \citep{ehsani2023deepshallow} led to:

\begin{equation}
E_{td}(T_h) = ConvLSTM(T_h) \text{ OR } DeepShallow(T_h)
\end{equation}

This particular layer deploys convolutional filters to detect spatial patterns within each time frame while simultaneously respecting the sequential nature of the data. Therefore, the temporal dependency encoder adeptly captures both the spatial and temporal aspects of the traffic data.

\subsubsection{3D Convolution Encoder}

To process fare closure data, the 3D convolution encoder (\( E_{3d} \)) applies volumetric filters, capturing relationships between data points in three dimensions:

\begin{equation}
E_{3d}(F_c) = BatchNorm(Conv3D(F_c))
\end{equation}

Batch normalization is employed here to maintain stability during training, normalizing across mini-batches.

\subsubsection{1D Seasonality Encoder}

We further incorporated an encoder (\( E_{1d} \)) designed to identify seasonality patterns. To maintain spatial relations and prevent network flattening, we employed upsampling techniques to amplify the dimensionality of seasonality through transpose convolution:
\begin{equation}
E_{1d}(S_f) = TransposeConv(Upsample(S_f))
\end{equation}

This technique prevents the loss of spatial relationships and supports the integrity of the seasonality signal.

\subsubsection{Decoder Layer}

The decoder layer (\( D \)) converges the outputs of the encoders into a comprehensive prediction of passenger traffic (\( P_t \)). It achieves this through:

\begin{equation}
P_t = CNN_{merge}\left(Concat\left[E_{td}(T_h), E_{3d}(F_c), E_{1d}(S_f)\right]\right) 
\end{equation}

In this layer, encoded features are concatenated and processed through convolutional layers (\( CNN_{merge} \)), which extract and combine high-level features. This streamlined process enables the decoder to analyze complex relationships within the data and output a precise, multidimensional traffic prediction for each flight leg, reflecting the nuanced interdependencies captured by the model.

% ---- REVISED UNTILL HERE DEC 18 2023 -----

\section{Results}

In this section, the results from various models used to predict flight-level passenger traffic are presented and discussed. The models range from the traditional ARIMA to more sophisticated deep learning structures, including a ConvLSTM network, both with and without spatial information preservation, and an advanced DeepShallow Network. These models are evaluated using  Mean Squared Error (MSE) loss, and the performance improvements are benchmarked against the baseline ConvLSTM model.

\autoref{tab:comparison}  summarizes the MSE losses and percentage improvements for each model, with the ConvLSTM with shared weights DeepShallow Network variant achieving the lowest MSE, indicating superior performance. The improvements in MSE loss are calculated relative to the baseline ConvLSTM model's performance.

\begin{table}[h!]
\centering
  \begin{tabular}{lcc}
    \toprule
    \textbf{Model} & \textbf{MSE Loss} & \textbf{Improvements (\%)} \\
    \midrule
    ARIMA & 6.121 & - \\
    SARMIA & 5.732 & - \\
    CNN & 5.069 & - \\
    ConvLSTM & 4.450 & - \\
    + Spatial & 3.011 & +32.33 \\
    + Shalow CNN & 2.973 & +33.18 \\
    + DeepShallow & 2.941 & +33.90 \\
    + Shared Weights & \textbf{2.934} & \textbf{+34.07} \\
  \bottomrule
\end{tabular}
\caption{Model Performance Comparison using MSE Loss for Predicting Flight-Level Passenger Traffic. The improvement percentages reflect the enhanced accuracy over the baseline ConvLSTM model. These statistics are compiled from an extensive dataset encompassing 50 diverse markets, with variations in capacity (from small to high-capacity markets), operational nature (covering both domestic and international flights), and network connectivity (including hub-to-hub, hub-to-spoke, and spoke-to-spoke connections).}
    \label{tab:comparison}
\end{table}

The deep learning models demonstrate a marked improvement over the ARIMA model, validating their enhanced capability to capture complex data patterns. Notably, the integration of spatial information into the ConvLSTM model significantly boosts performance, with each additional feature leading to further improvements. The DeepShallow Network with shared weights showcases the most substantial performance gain, underscoring the benefits of capturing both spatial and temporal dynamics and leveraging shared weights for improved generalization. Future work includes further analysis to assess the statistical significance of these results and additional visualizations for a more comprehensive evaluation.

For a more comprehensive understanding of the models' performances and an evaluation of their specific strengths and weaknesses, further analysis and comparison based on different metrics are presented in the analysis section (\autoref{sec:analysis}).

\section{Analysis}
\label{sec:analysis}

In this section, we delve into a multifaceted analysis of our model, scrutinizing its design and operational efficacy. This rigorous evaluation includes optimizing hyperparameters to enhance model performance, dissecting the model's proficiency in capturing seasonal fluctuations and long-term trends in passenger traffic, and conducting a sensitivity analysis to gauge the model's resilience to anomalous data variations. Such a detailed exploration is instrumental in delineating the model's capabilities, pinpointing potential areas for refinement, and steering further development.

\subsection{Hyperparameter Tuning}

The optimization of hyperparameters is vital in our modeling process, focusing on two essential aspects: optimal window size for historical data and neural network parameter tuning. Window size reflects the volume of historical data used for predictions. An inadequate window can neglect critical trends, whereas an excessively large one may complicate the model and invite overfitting. We investigated varying window sizes to find an equilibrium that maximizes predictive accuracy without overcomplicating the model.

For the neural network configuration, we fine-tuned parameters including layer quantity, filter count, kernel dimensions, and initial layer settings in the DeepShallow architecture. Beginning with a random search, we identified key hyperparameters rapidly, which, supplemented by Bayesian optimization \citep{turner2021bayesian}, refined the values to optimal levels.

\begin{figure}[h!]
    \centering
 \includegraphics[width=\linewidth]
    {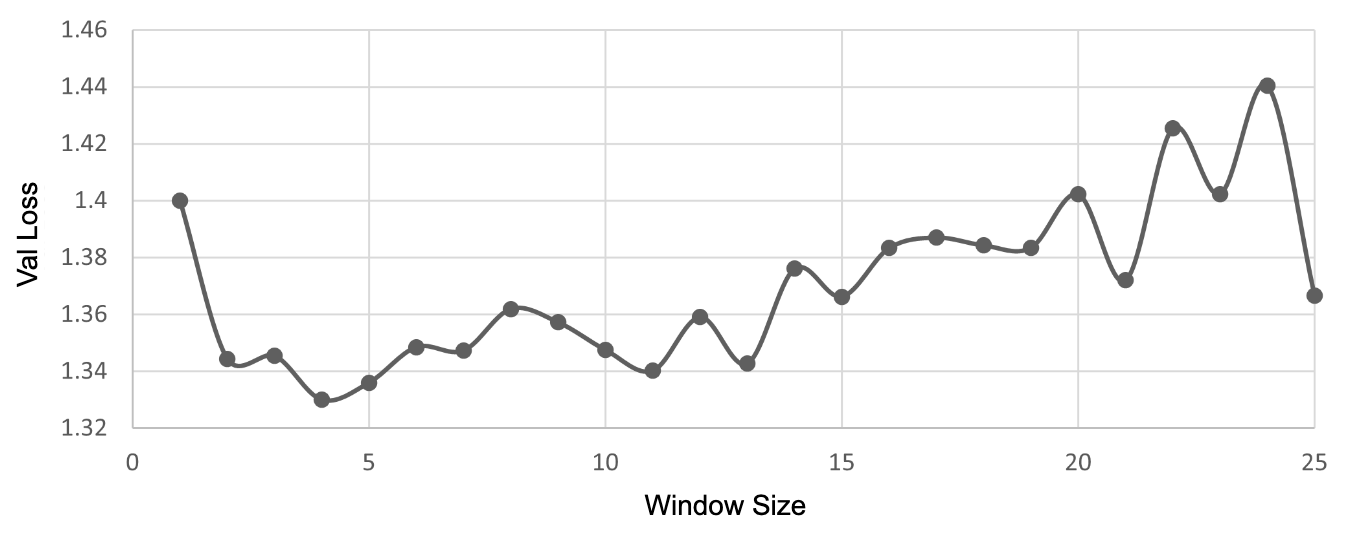}
    \caption{Validation Loss Variation with Window Size for the DeepShallow Network Model. This graph illustrates how different historical window sizes influence the model's validation loss, with the trend indicating the optimal range for balancing model complexity and predictive performance. The data points represent aggregated results across a spectrum of market conditions.}
    \label{fig:window_size}
\end{figure}

Through careful hyperparameter tuning, we ensured that our model was both robust and accurate. For instance, our investigation into window size revealed that a window size of approximately five led to a noticeable increase in performance, with larger sizes resulting in diminishing returns and a subsequent drop in performance. The effects of window size on validation set MSE loss, as shown in \autoref{fig:window_size}, reinforce the benefit of having a view of previous flights' traffic data.

\subsection{Seasonality and Trend Analysis}

A crucial advantage of our methodology lies in its capacity to accurately capture both the seasonality and trends inherent in flight-level passenger traffic data. This enhanced capability is evident when evaluating day-to-day predictions on our test set across various markets, in which the performances of different models were comparatively assessed.

Our method has shown a significant proficiency in predicting data points closer to the set reference point, i.e., nearer to the departure date. In addition, it also maintained an appreciable consistency in tracking the traffic trend for data points further from the reference point, i.e., further from departure. Moreover, our model has displayed a strong ability to recognize seasonality, effectively capturing the variations in traffic across different seasons. This is largely attributable to the incorporation of the seasonality data component into our model's training data, allowing the model to account for these cyclic variations.

\begin{figure}[h!]
\centering
\includegraphics[width=\linewidth]
{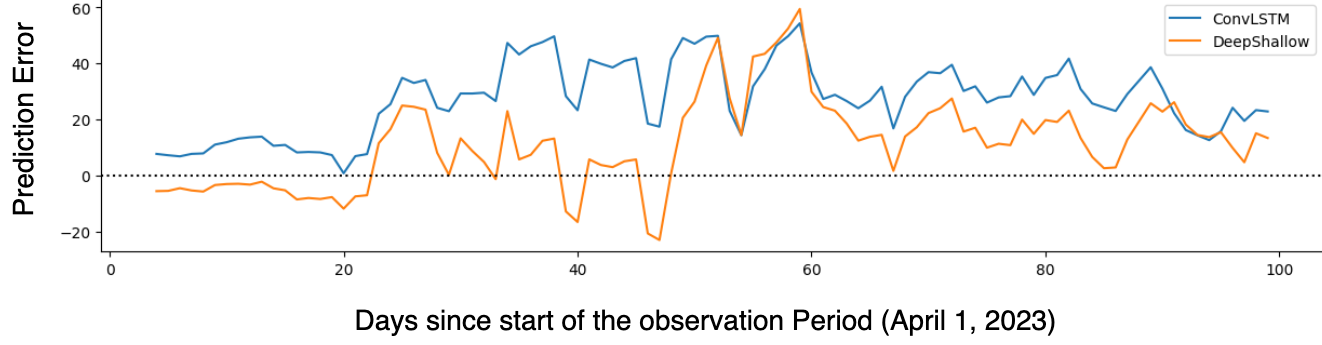}
\caption{Performance Comparison Between ConvLSTM+Spatial and DeepShallow Models Over a 100-Day Period. This graph depicts the absolute differences in predictions from the actual observed traffic data, highlighting the prediction accuracy for each model. The data, averaging results from 12 diverse markets, illustrates the temporal prediction trend from April 1, 2023, to July 10, 2023. The DeepShallow network’s curve demonstrates its relative prediction performance against the ConvLSTM+Spatial model across the observed period.}
\label{fig:seasonality_trend_analysis}
\end{figure}

Quantitative analyses, detailed in \autoref{fig:seasonality_trend_analysis}, reveal our model’s capability in adapting to seasonal shifts, outperforming traditional models that often fail to accommodate such cyclicality. This adaptability is crucial for strategic planning in the aviation industry, where understanding and anticipating passenger flow is vital \citep{zachariah2023systematic}.

\autoref{fig:seasonality_trend_analysis} presents a comparative analysis of the absolute discrepancies between predictions made by ConvLSTM+Spatial, the DeepShallow network, and the actual traffic. The x-axis represents the departure dates, starting from our reference date (April 1, 2023), spanning across a 100-day interval. The y-axis shows the absolute difference between the predicted and observed values, highlighting the model's over and under predicting. The DeepShallow network consistently outperforms the ConvLSTM+Spatial, indicating its superior capability in both capturing the inherent trends and adjusting for seasonality in the flight-level passenger traffic data. This comparative analysis underscores the robust performance of our method in identifying and adapting to the seasonal and trend-based variations in flight traffic data.

%   Add a figure showing the result of the DeepSHallow vs baselines vs actual based on a reference point

\subsection{Sensitivity Analysis}
\label{Sensitivity_Analysis}

Given the inherently unpredictable nature of real-world events, it is essential to assess the robustness of our model under various adverse scenarios. This involves conducting a sensitivity analysis to evaluate how changes in certain parameters influence the model's predictive performance.

One aspect of this analysis is the temporal sensitivity of the model, i.e., how the model performs on specific dates, including those marked by extraordinary events that could significantly deviate from the usual traffic trends. Another consideration is the model's sensitivity to changes in the flight specifics, such as a switch in the aircraft type, which could directly impact the available seats and thus the passenger traffic.

\begin{figure}[h!]
\centering
\includegraphics[width=\linewidth]
{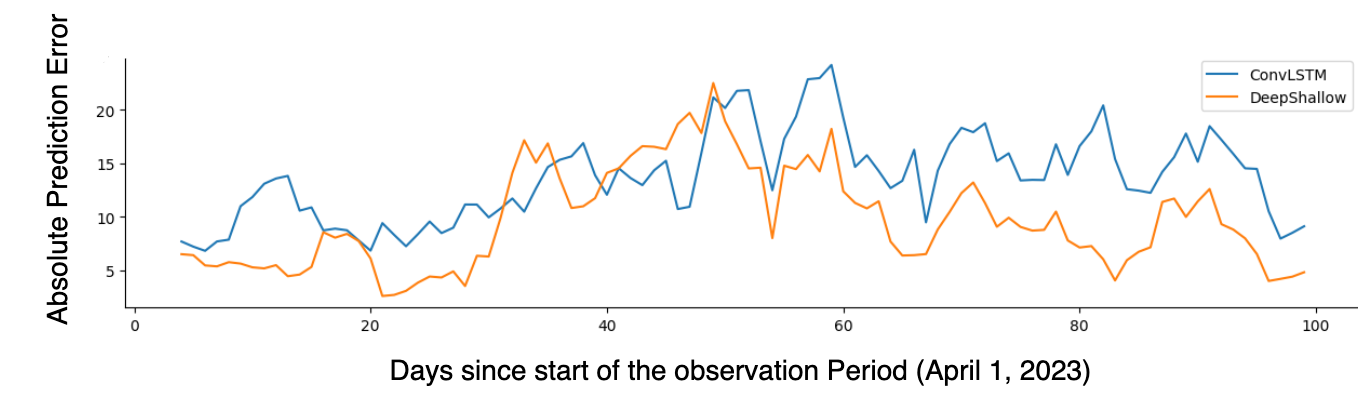}
\caption{Differential Performance in Sensitivity Analysis for ConvLSTM+Spatial vs. DeepShallow Network. The chart illustrates the deviation between actual and predicted passenger traffic over a 100-day period, with the x-axis representing days since April 1, 2023, and the y-axis showing the prediction error. Positive values along the y-axis indicate instances where the model overestimated traffic, while negative values indicate underestimation. A pronounced change in the prediction error on the 20th day corresponds to an imposed scenario of increased airplane size, testing the models' adaptability to sudden shifts in traffic patterns.}

\label{fig:sensitivity_analysis}
\end{figure}

\autoref{fig:sensitivity_analysis} presents a comparative sensitivity analysis by delineating the differential performance of ConvLSTM+Spatial and DeepShallow network. Here, we examine the model's reaction to an increase in airplane size leading to elevated traffic, introduced on the 20th day from the reference date. The x-axis displays the departure dates over a 100-day interval beginning from our reference date of April 1, 2023. The y-axis exhibits the differential performance, calculated as the difference between observed and predicted traffic values. This differential metric provides insight into the model's propensity to over-predict (resulting in positive values) or under-predict (yielding negative values) the passenger traffic.

The results underscore the adaptability of the DeepShallow network relative to the ConvLSTM+Spatial model, demonstrating the ability to swiftly adjust its predictions in response to the surge in traffic precipitated by the change in airplane size, thereby effectively capturing the underlying traffic trend. This insight signifies the utility of the shallower component of the DeepShallow network in enhancing the model's adaptability.

\section{Conclusion}

This study marks a significant advancement in flight-level passenger traffic prediction, with the introduction of advanced deep learning techniques that have shown to significantly enhance traffic prediction accuracy. By embedding 3D data transformation, multimodal pattern detection, and innovative data strategies into our deep learning network, we've not only achieved substantial performance gains over both classical and contemporary models but also laid a groundwork for more dynamic and robust airline pricing and management. Empirically, our model has demonstrated a remarkable 70\% improvement over classical forecasting methods and a 34\% enhancement compared to standard deep learning approaches. While the model requires greater computational resources and has yet to be tested on a global scale, the improvements observed promise considerable impact on the operational efficiency and strategic adaptability in the airline industry.

\section{Future Work}

Future research initiatives will aim to scale our proposed model to more extensive airline datasets, transcending the current market-by-market analysis. Applying the model to a network-wide dataset promises to shed light on broader passenger demand trends, though it also introduces the complexity of diverse market characteristics.

To navigate this complexity, we propose leveraging unsupervised learning techniques \citep{wang2020application}, such as an unsupervised masking strategy inspired by Masked Language Modeling in NLP \citep{devlin2018bert}. By intentionally obscuring parts of the input data, the model would be challenged to infer missing values, thereby gaining deeper insights into the underlying patterns of airline traffic.

Subsequent stages would involve fine-tuning this pre-trained model to adapt to specific market clusters or individual markets, leveraging its generalized knowledge while catering to unique market features. Through this approach, we anticipate that the model will achieve a fine balance between general applicability and market-specific accuracy, leading to improved capabilities in forecasting passenger demand. Such advancements could prove invaluable for strategic airline management, offering a promising avenue for future research.

\bibliographystyle{ieeetr}
\bibliography{conference_101719}

%% Ke, Jintao, et al. “Short-Term Forecasting of Passenger Demand under on-Demand Ride Services: A Spatio-Temporal Deep Learning Approach.” arXiv.Org, 20 June 2017, arxiv.org/abs/1706.06279. 

%% Bell, Franziska, and Slawek Smyl. “Forecasting at Uber: An Introduction | Uber Blog.” Forecasting at Uber: An Introduction, 6 Sept. 2018, www.uber.com/blog/forecasting-introduction/. 

\end{document}